# Alignment among Language, Vision and Action Representations

**Nicola Milano[1,2*] and Stefano Nolfi[1]**

[1] Institute of Cognitive Sciences and Technologies, National Research Council,

Roma, Italy

[2] University of Naples "Federico II", Natural and Artificial Cognition Laboratory "Orazio Miglino",

Napoli, Italy

*corresponding author: nicola.milano@unina.it

**Abstract**

*A fundamental question in cognitive science and AI concerns whether different learning modalities: language, vision, and action, give rise to distinct or shared internal representations. Traditional views assume that models trained on different data types develop specialized, non-transferable representations. However, recent evidence suggests unexpected convergence: models optimized for distinct tasks may develop similar representational geometries.*
*We investigate whether this convergence extends to embodied action learning by training a transformer-based agent to execute goal-directed behaviors in response to natural language instructions. Using behavioral cloning on the BabyAI platform, we generated action-grounded language embeddings shaped exclusively by sensorimotor control requirements. We then compared these representations with those extracted from state-of-the-art large language models (LLaMA, Qwen, DeepSeek, BERT) and vision-language models (CLIP, BLIP).*
*Despite substantial differences in training data, modality, and objectives, we observed robust cross-modal alignment. Action representations aligned strongly with decoder-only language models and BLIP (precision@15: 0.70–0.73), approaching the alignment observed among language models themselves. Alignment with CLIP and BERT was significantly weaker. These findings indicate that linguistic, visual, and action representations converge toward partially shared semantic structures, supporting modality-independent semantic organization and highlighting potential for cross-domain transfer in embodied AI systems.*

## 1. Introduction

Internal representations in neural networks mediate the transformation from input to output data. It is therefore commonly assumed that networks exposed to different data modalities or optimized for distinct tasks will develop correspondingly specialized representational structures. This assumption stems from the idea that the internal feature spaces of neural models are shaped by both the statistical properties of their training data and the constraints imposed by their learning objectives. Consequently, variations in data modality or task demands are expected to yield task-dependent, non-transferable representations rather than shared or generalizable ones.

This view has particularly strong implications for understanding the nature of semantic knowledge. Traditionally, it has led to the belief that systems trained solely on linguistic input cannot acquire grounded or embodied representations of the kind that emerge from direct perceptual and motor interaction with the physical world. In this perspective, language-based models operate within a self-contained symbolic domain, lacking the experiential substrate necessary for forming concepts anchored in sensorimotor experience. Without explicit grounding, language models are thought to remain limited to syntactic or statistical correlations among symbols, unable to capture the rich conceptual knowledge that humans acquire through interaction with their environment (for a discussion see Bender & Koller, 2020; Sahlgren & Carlsson, 2021; Piantadosi & Hill, 2022; Mitchell & Krakauer, 2023). Under this framework, meaningful semantic representations would necessarily be modality-specific: linguistic knowledge would remain fundamentally separate from visual understanding, and both would be categorically distinct from the action-grounded knowledge embodied agents acquire through sensorimotor experience.

If this traditional view were correct, we would expect minimal structural correspondence among representations learned through different modalities. Language models trained on text corpora, vision-language models trained on image-caption pairs, and embodied agents trained through physical interaction should develop fundamentally incompatible representational geometries, each shaped exclusively by its unique training regime. However, this theoretical prediction is being increasingly challenged by empirical observations.

Recent empirical findings challenge this traditional boundary, revealing that systems trained on data differing in modality, content, or task often converge toward similar internal representational geometries. Evidence also suggests that large language models (LLMs) acquire implicit knowledge about the physical world from textual data alone and can generalize or reason about non-linguistic phenomena, raising questions about the necessity of direct sensorimotor grounding for semantic representations.

For example, Merullo et al. (2023) demonstrated that the conceptual representations learned by LLMs trained on next-word prediction tasks are functionally similar to those acquired by vision–language models (VLMs) trained on image–text alignment tasks. Specifically, the representations of frozen text-only LLMs can be linearly mapped onto those of VLMs, indicating a substantial overlap in their underlying representational structures. Similarly, Li et al. (2024) found that LLMs and VLMs partially converge toward isomorphic embedding spaces, further suggesting cross-modal alignment.

Complementary findings reveal that LLMs trained exclusively on text exhibit forms of grounded understanding. For instance, such models (i) learn color representations consistent with human perceptual judgments (Abdou et al., 2021; Patel & Pavlick, 2022; Søgaard, 2023); (ii) construct and update spatial representations of described environments as narratives unfold (Li et al., 2021; Patel & Pavlick, 2022; Bubeck et al., 2023); and (iii) infer affordances, identifying which actions an agent can or cannot perform in a given context (Jones et al., 2022), (iv) generate action plans (Huang et al., 2022). These observations collectively suggest that linguistic co-occurrence statistics encode substantial information about the physical and functional structure of the world, sufficient for models to approximate aspects of grounded cognition.

Building on these insights, the present study investigates a critical gap in our understanding of cross-modal representational alignment: whether the internal representations learned through embodied action, which directly mediate goal-directed behavior in situated agents, align with those learned through passive observation of language and vision. This question is fundamental because action representations constitute perhaps the most grounded form of knowledge, shaped not by statistical patterns in observed data but by the pragmatic demands of achieving goals through physical interaction. If action representations align with those of language and vision models, it would provide the strongest evidence yet that core semantic structures transcend modality.

Specifically, we ask: Do the representations acquired by LLMs and VLMs—systems that learn through passive statistical analysis of text and images—align with those learned by a language-conditioned robotic agent trained to perform actions that achieve goals expressed in natural language? We hypothesize that if semantic organization is fundamentally modality-independent, then substantial alignment should exist even between passive linguistic/visual learning and active sensorimotor learning.

The novelty of this work lies in two aspects. First, it extends prior analyses of cross-modal alignment by including action representations, i.e., the internal structures that mediate the generation of goal-directed behavior rather than merely predicting or classifying perceptual inputs. Second, it contrasts learning through passive observation (as in LLMs and VLMs which process pre-existing text and image data) with learning through embodied interaction (as in language-conditioned robots, where the agent actively shapes its sensory experiences through its own actions and learns from the consequences of those actions).

To address this question, we trained a transformer-based model on the BabyAI platform (Chevalier-Boisvert et al., 2018) using behavioral cloning, enabling a simulated agent to execute action sequences in response to natural language instructions. We then compared the resulting action representations with those generated by state-of-the-art LLMs (LLaMA, Qwen, Deepseek, BERT) and VLMs (CLIP, BLIP).

Our findings reveal robust alignment between action-grounded representations and those learned by decoder-only language models and BLIP, with precision@15 scores (0.70–0.73) approaching the alignment observed among language models themselves. This convergence occurs despite the models being trained on fundamentally different modalities and objectives, i.e. text prediction, image-text association, and goal-directed action control. Such alignment suggests the existence of shared semantic structures that transcend modality, linking linguistic, visual, and sensorimotor domains within a unified representational framework. These results challenge the traditional view that grounded knowledge requires direct sensorimotor experience and suggest that language, vision, and action may converge on common representational principles that enable cross-domain transfer and integration.

## 2. Method

In this section we describe the BabyAI platform, the architecture of the transformer-based neural network controller, the LLMs and VLMs used to generate alternative representations, and the methodology used to measure the alignment among representations extracted by different models.

### 2.1 The BabyAI platform

The BabyAI research platform (Chevalier-Boisvert et al., 2018) is a Python library designed to investigate language-conditioned action learning—that is, the acquisition of capabilities enabling a robotic agent situated in an external environment to respond to natural language requests, such as "put a ball next to the blue door," by generating action sequences that achieve the specified goal.

The platform operates within a 2D, partially observable environment populated with objects including balls, boxes, keys, and doors in various colors: yellow, green, blue, red, grey, and purple (see Figure 1). At the beginning of each evaluation episode, the agent's position, orientation, and the types and locations of objects in the environment are randomly initialized.

The agent can execute six actions: "turn left," "turn right," "go forward," "pick up," "drop," and "open." These actions enable the agent to complete a range of missions described through natural language instructions. Instructions comprise phrases such as "go to," "pick up," "put next to," and "open door," combined with object descriptions, colors, and connecting words such as "and," "then," "after," "right," and "left," which encode temporal and spatial relationships.

Missions are categorized into task types requiring similar behaviors. For instance, GO-TO tasks involve locating the correct object, navigating toward it while avoiding other objects, and approaching it. The precise action sequence required to fulfill a language instruction depends on: (i) the instruction itself, (ii) the agent's initial position and orientation, (iii) the objects present in the environment, and (iv) their locations.

Agents receive two inputs during operation: the language instruction, which remains constant throughout each episode, and an image displaying the local portion of the environment visible from the agent's current position, which varies at each step as the agent moves, turns, and manipulates the environment (Figure 1). The initial position and orientation of the agent, the objects present in the environment, and their locations are randomly determined at the beginning of each evaluation episode. The language request for each episode is randomly selected from the complete set of requests that can be satisfied in the given environment.

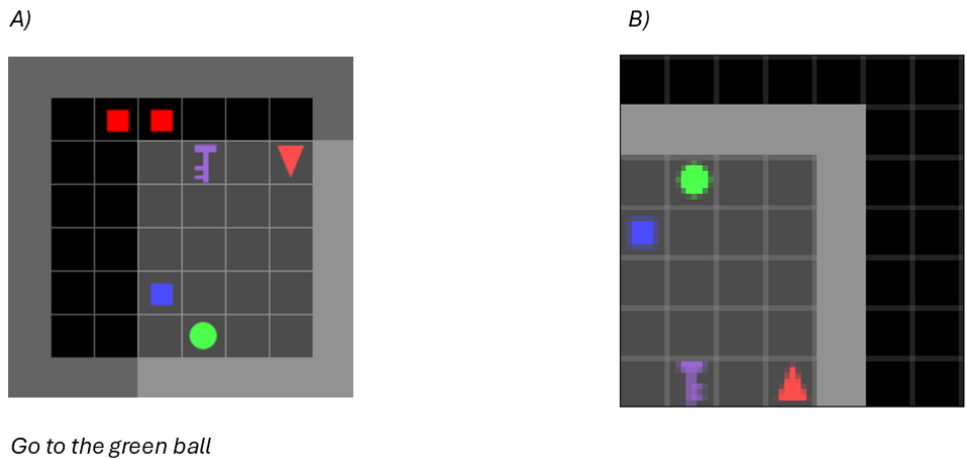

*Go to the green ball*

Figure 1: Illustration of the BabyAI platform. Panel A displays an example environment consisting of a single room with a 6×6 grid. The red triangle, filled circle, filled square, empty square with a line, and key symbol represent the agent, ball, boxes, door, and key objects, respectively. The text at the bottom shows an example language request that can be encountered in this environment: "Go to the green ball." Panel B displays the local portion of the environment perceived by the agent from its current position and orientation.

In our study, we consider three tasks: **GO-TO**, which requires the agent to navigate toward a specified target object (e.g., "go to the red ball"); **PICK-UP**, which requires the agent to approach and retrieve a designated object (e.g., "pick up the blue key"); and **ESCAPE-FROM**, which requires the agent to move away from a designated object (e.g., "escape from the blue box"). In addition, we introduce synonymous expressions to allow the same action to be described using alternative linguistic instructions. The vocabulary employed in the experiments is detailed in Section 2.2.

## 2.2 The transformer-based neural network controller

The neural network model used to solve the BabyAI task is a visuo-linguistic architecture in which the visual and linguistic components are processed separately to produce embedded representations of the input image and the language instruction. These representations are subsequently integrated to generate the action output. The language request is represented as a sequence of token embeddings, where each token corresponds to a word or sub-word unit in the input sentence. Initially, each token is assigned an integer index ranging from 0 to V, where V denotes the vocabulary size. Each index is then mapped to a 100-dimensional embedding vector, forming the initial token representations. These embeddings are initialized as identical vectors but are subsequently differentiated during the learning process, enabling them to capture semantic and syntactic distinctions between tokens. The sequence of token embeddings is processed by a transformer block with multi-head self-attention followed by a multi-layered-perceptron that transforms word level embeddings into a single 128-

dimensional embedding vector $E_L$, effectively summarizing the entire input sentence in a compact numerical representation suitable for downstream processing (Figure 2, bottom right).

The images are mapped into embedding vectors $E_V$ through a residual convolutional neural network (ResNet CNN) that extracts multiscale visual features at various resolutions (Figure 2, bottom left). This network is pre-trained on an image classification task using an existing image database with associated labels. Specifically, it comprises two convolutional ResNet layers pre-trained on ImageNet (Deng et al., 2009; He et al., 2016), along with two additional untrained convolutional layers.

The $E_L$ and $E_V$ vectors are subsequently integrated through a Language-Attends-to-Vision transformer block. In this block, cross-attention is performed with language serving as the query and flattened multiscale visual tokens acting as keys and values. The resulting output is fed into a deep residual multi-layer perceptron (MLP). This MLP encodes the predicted next action using six neurons, corresponding to the six possible actions that BabyAI agents can execute (see Figure 2).

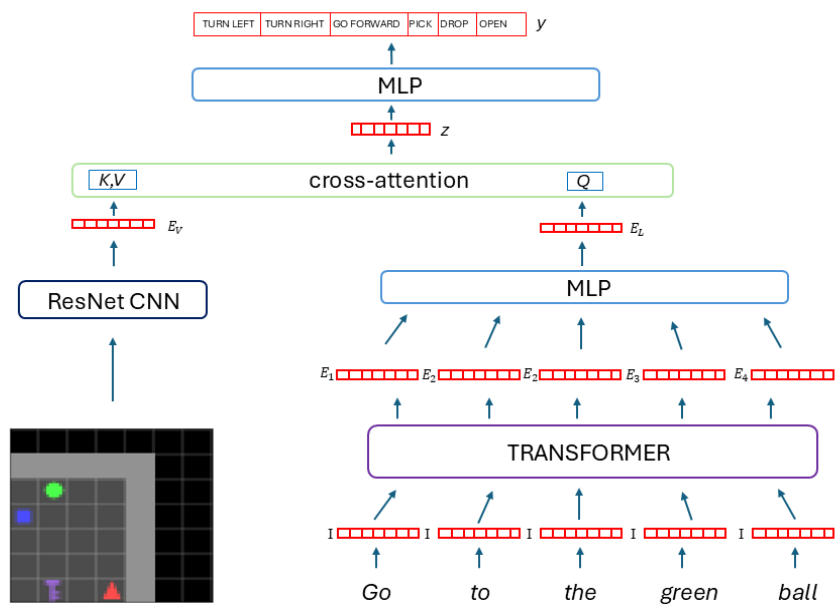

Figure 2. Schematic depiction of the neural architecture of the agent.

Overall, the architecture and training process can be described by the following equations:

$$\boldsymbol{E}_V = f_{vis}(O_t) \tag{1}$$

$$\boldsymbol{E}_L = f_{lang}(L) \tag{2}$$

$$\boldsymbol{z}_t = CMA(\boldsymbol{E}_L, \boldsymbol{E}_v) \tag{3}$$

$$\boldsymbol{z}'_t = MLP(\boldsymbol{z}_t) + \boldsymbol{z}_t \tag{4}$$

$$a_t = softmax(\boldsymbol{z}'_t) \tag{5}$$

where $O_t$ denotes the RGB observation at time step $t$, $L$ is the language instruction, $f_{vis}$ is the visual encoder (ResNet CNN), and $f_{lang}$ maps the tokenized instruction to a sentence-level embedding as described earlier. The cross-modal attention (CMA) module uses the language embedding as query and the visual embedding as keys and values. The output is processed through a residual MLP and projected onto a six-dimensional action space, corresponding to the available actions in the BabyAI environment.

The model is trained by using a behavioral cloning (BC) method, which involves minimizing the cross-entropy between the actions produced by the network and the actions generated by the demonstrator. The training set is produced by:

- Generating all possible well-formed language requests that can be created using the following vocabulary: "go to", "move towards", "pick up", "escape from", "stay away", "avoid", "the", "box", "key", "ball", "red", "green", "grey", "yellow", "purple", "blue", resulting in a total of 108 distinct language requests.
- Generating multiple demonstrations episodes for each language request up to a total of 10,000 demonstration episodes. At the start of each demonstration, the agent's initial position and orientation, as well as the type and position of the objects in the environment, are randomly assigned. The sequence of observations and actions executed by the agent is generated using the demonstrator bot included in the BabyAI platform, which is capable of controlling the agent to satisfy any possible language request.

The training data of each demonstration episode includes the following elements: (i) a language instruction (*L*), (ii) the RGB image experienced by the agent as its perceived environment (*O*), and (iii) the sequence of actions chosen by the bot (*a*). The visual input consists of 64x64 pixel RGB images. The training set comprises a vector of triplets <$O_{ij}, L_{ij}, a_{ij}$>; where *i* represent the episode index (ranging from 1 to N), and *j* denotes the step index (ranging from 1 to S). The policy is trained to produce the actions $a_{ij}$ by receiving the input pair <$O_{ij}, L_{ij}$>.

In our model, the embedding vectors of words token are initialized as identity vectors ***I,*** and the network that maps the word embeddings into the $\boldsymbol{E}_L$ embedding vector is initialized with random weights. This ensures that the embeddings representing both individual words and the overall language request are shaped entirely through action learning, i.e. exclusively via the behavioral cloning process.

As a result, the model develops representations that support only the generation of actions. Our dynamical learning of language embeddings fundamentally differs from previous works that relies on pre-trained visuo-language model to generate the embeddings for the language instruction (Lynch et a., 2023; Milano and Nolfi, 2025).

## 2.4 Reference models

We refer to the $E_L$ embeddings generated by our model from language requests as *action-language embeddings*, as they are specifically shaped to support the generation of behaviors that satisfy verbally specified goals. Accordingly, we refer to our model trained via behavioral cloning to map language instructions to motor behaviors as *action language model* (ALM).

Our objective is to quantify the correlation between the vector space induced by the ALM and those of VLMs and LLMs. More precisely, we aim to compare embeddings learned through the mapping of language instructions to actions with embeddings derived from models that capture regularities in multimodal vision-language data or statistical regularities in large-scale language corpora.

For this purpose, we consider embeddings generated through the CLIP and BLIP VLMs and through the Llama, Deepseek, and Qwen LLMs. CLIP and BLIP naturally produce a single embedding vector for each input sentence. LLMs, in contrast, produce contextualized embedding vectors for each token. In these models, a single embedding vector can be obtained by averaging the embedding vectors of the last stacked layer in encoder-only models such as BERT or of the last layer preceding the output layer in decoder-only models such as Llama, Deepseek, and Qwen.

The **CLIP** (Contrastive Language–Image Pre-training) model, introduced by Radford et al. (2021), is a multimodal neural network designed to learn joint representations of images and text. Its primary function is to align visual and linguistic information so that it can understand and relate images to their corresponding textual descriptions without requiring task-specific fine-tuning. CLIP is trained using a contrastive learning objective: given a large set of image–text pairs, the model learns to associate each image with its correct caption while distinguishing it from mismatched ones. This is achieved by encoding images and text separately—using a vision transformer (ViT) or convolutional network for images and a transformer-based language model for text—and maximizing the cosine similarity between the embeddings of matching pairs. The training dataset, known as WebImageText (WIT), consists of approximately 400 million image–text pairs collected from the internet, making it one of the largest and most diverse multimodal datasets available at the time.

Building upon the multimodal foundation established by CLIP, the **BLIP** (Bootstrapped Language–Image Pre-training) model (Li et al., 2022) further advances vision–language understanding by integrating both contrastive and generative learning objectives. While CLIP focuses primarily on aligning image and text embeddings, BLIP is designed to perform a broader range of tasks, including image–text retrieval, caption generation, and visual question answering. It employs a unified encoder–decoder architecture that can switch between vision–language understanding and generation modes. BLIP is trained using a combination of noisy web data and curated datasets, but introduces a bootstrapped captioning approach that refines the quality of textual data: the model first generates improved captions for noisy image–text pairs and then retrains on this cleaner data. This iterative process enhances the alignment between visual and textual modalities, allowing BLIP to achieve state-of-the-art performance on several benchmarks.

**BERT** (Bidirectional Encoder Representations from Transformers; Devlin et al., 2019) is built on an encoder-only transformer architecture and processes words in relation to all surrounding words simultaneously. It is pre-trained using two key objectives: masked language modeling (MLM) and next sentence prediction (NSP). In MLM, a percentage of input tokens are randomly masked, and the model is trained to predict these masked tokens based on their context. NSP involves predicting whether one sentence logically follows another, helping the model learn sentence relationships. BERT uses the [CLS] special token to indicate the beginning of a sentence and the [SEP] special token to separate sentences. The model was pre-trained on a corpus of approximately 3.3 billion words, comprising the BooksCorpus (800 million words) and English Wikipedia (2.5 billion words).

**LLaMA** (Large Language Model Meta AI; Touvron et al., 2023) is an autoregressive transformer-based language model developed by Meta, based on a decoder-only architecture. It is trained using standard causal language modeling, where the model learns to predict the next token in a sequence given the preceding context. The training corpus consists of a diverse and carefully curated mixture of publicly available datasets, including academic publications, web content, and other high-quality sources. The total size of the training data reaches approximately 1.4 trillion tokens.

**Qwen** (Qwen Language Model, Alibaba DAMO Academy; Bai et al., 2023) is a decoder-only autoregressive transformer-based language model developed by Alibaba. It is trained on standard causal language modeling, where the model learns to predict the next token in a sequence based on the preceding context. The training corpus includes a diverse and extensive collection of high-quality multilingual and domain-specific data, such as web documents, technical content, and code. The total size of the training data exceeds 2.2 trillion tokens.

**DeepSeek** (DeepSeek AI; Guo et al., 2025) is also a decoder-only autoregressive transformer-based language model trained with causal language modeling. The training dataset comprises 800 billion tokens from a diverse collection of publicly available text sources, including web data, books, and domain-specific corpora.

**2.5 Extraction of embeddings**

In this section we describe the method used to generate the embeddings with alternative models

For our action language model (ALM), embeddings are obtained from the multi-layer network shown in the bottom-right portion of Figure 2. This network maps the sequence of words forming each language instruction into a single 128-dimensional vector representation. Word embeddings are initialized as identical vectors and are subsequently shaped during training through behavioral cloning to support action generation. As a result, the resulting action-language embeddings are exclusively optimized to facilitate the production of goal-directed behaviors.

For the CLIP and BLIP vision–language models (VLMs), embeddings are extracted from the text encoder component. These models consist of separate text and visual encoders that process sentences and images, respectively, producing corresponding embedding vectors. Visual and text encoders are trained using contrastive learning objectives that encourage embeddings of matched image–text pairs to be similar, while pushing apart embeddings of mismatched pairs. Consequently, the sentence embeddings produced by these models jointly reflect linguistic and visual information.

For decoder-only large language models (LLaMA, Qwen, and DeepSeek), embeddings are extracted from the final hidden layer immediately preceding the output layer used for next-token prediction. These transformer-based models generate contextualized embeddings for each token in the input sequence. To obtain a single sentence-level representation, we average the contextualized token embeddings from the final layer.

For BERT, an encoder-only language model, the sentence embedding is extracted from the contextualized representation of the final layer corresponding to the special [CLS] token, which is conventionally used as a summary representation of the entire input sentence. The embeddings of language models are thus shaped by the need to predict the next token, in the case of decoder-only models, or to infer missing tokens, in the case of encoder only models.

Across all models, the same set of sentences—corresponding to the 108 language instructions available in the BabyAI platform—was used for embedding extraction. No additional prompts, prefixes, or formatting were introduced. None of the VLMs or LLMs were fine-tuned for this study; thus, the extracted embeddings reflect only the representations learned from the original models.

Finally, when comparing embedding spaces with different dimensionalities, we applied principal component analysis (PCA) to reduce the higher-dimensional space to match the dimensionality of the lower-dimensional one prior to Procrustes analysis, as described in Section 2.6.

**2.6 Correlation measures**

To assess the similarity between the vector spaces produced by our action model, vision–language models (VLMs), and large language models (LLMs), we used the 108 sentences that can serve as language instructions in the BabyAI platform. For each model, we first computed a $108 \times 108$ similarity matrix containing the pairwise cosine similarities between sentence embeddings within that vector space. We then evaluated the alignment between these similarity matrices across models.

To quantify this alignment, we employed the precision-at-k (P@k) metric (Conneau et al., 2018), which is commonly used to assess the correspondence between vector spaces, for example, when evaluating the consistency of multilingual word embeddings produced by a single model. This approach involves identifying the k nearest neighbors of each embedding based on cosine similarity and compute the proportion of overlapping neighbors between two embedding spaces. Formally,

$$P@k = \frac{|listM_k \cap listN_k|}{k} \qquad (6)$$

where $listM_k$ and $listN_k$ denote the top-*k* nearest neighbors in the embedding spaces $M$ and $N$, respectively. Averaging Equation (1) across all embeddings yields the mean P@k for the set of sentences. In our analysis, *k* was varied from 1 to 15. P@k ranges between 0 and 1, with higher values indicating greater structural similarity between embedding spaces and lower values indicating weaker alignment.

This metric is more conservative than other measures commonly used for similar analyses, such as pairwise matching accuracy, percentile rank, or Pearson correlation (Minnema & Herbelot, 2019; Li et al., 2024).

To further assess the extent to which action, vision, and language representations are isomorphic, we applied Procrustes analysis to measure shape similarity by optimally aligning the representations of the behavioral cloning model with those of the VLMs and LLMs, using the same set of 108 sentences. Given the behavioral cloning embedding matrix $A$ (action representations of language instructions) and the language or visuo-language embedding matrix $B$, we employed orthogonal Procrustes analysis to estimate the transformation matrix $\Omega$ that best maps $A$ onto $B$. Specifically, we solved the constrained optimization problem:

$$\Omega = \arg\min_{R} \quad \| RA - B \|^2 \text{ s.t. } R^\top R = I, \tag{7}$$

which admits a closed-form solution via singular value decomposition (SVD). Letting $U\Sigma V^\top = \text{SVD}(BA^\top)$; the optimal mapping is given by $\Omega = UV^\top$. The aligned representations are then $\hat{A} = \Omega A$.

We quantify the disparity between the action and visuo-language spaces by the residual sum of squared distances after alignment:

$$D = \| \hat{A} - B \|^2, \tag{8}$$

which serves as a scale-independent metric of mismatch between modalities. Lower values of $D$ indicate closer correspondence between $A$ and $B$, whereas higher values reflect greater structural divergence.

Because Procrustes analysis requires the source and target spaces to have the same dimensionality, we apply principal component analysis (PCA) to reduce the dimensionality of the higher-dimensional space whenever a mismatch occurs.

## 3. Results

We trained 10 language-action networks initialized with randomly different connection weights. By analyzing the performance after 20,000 training epochs, we observed that the networks solve the ESCAPE task up to 95% of the cases and the GO-TO and PICK-UP tasks in 85% of the cases, independently from the specific words used to formulate the instructions.

Because we are interested in how the emergent embedding space of our model compare to those of pretrained language and vision-language models, we used the trained model to generate 108 action embedding vectors. We then computed the cosine-similarity matrix of these embeddings and compared the resulting P@k scores with those derived from the other models' similarity matrices to assess the degree of similarity between embedding spaces.

Figure 3 reports the P@15 alignment (Conneau et al., 2018) among the $108 \times 108$ cosine similarity matrices derived from the seven models as well as the similarity matrix obtained from randomly generated embedding vectors.

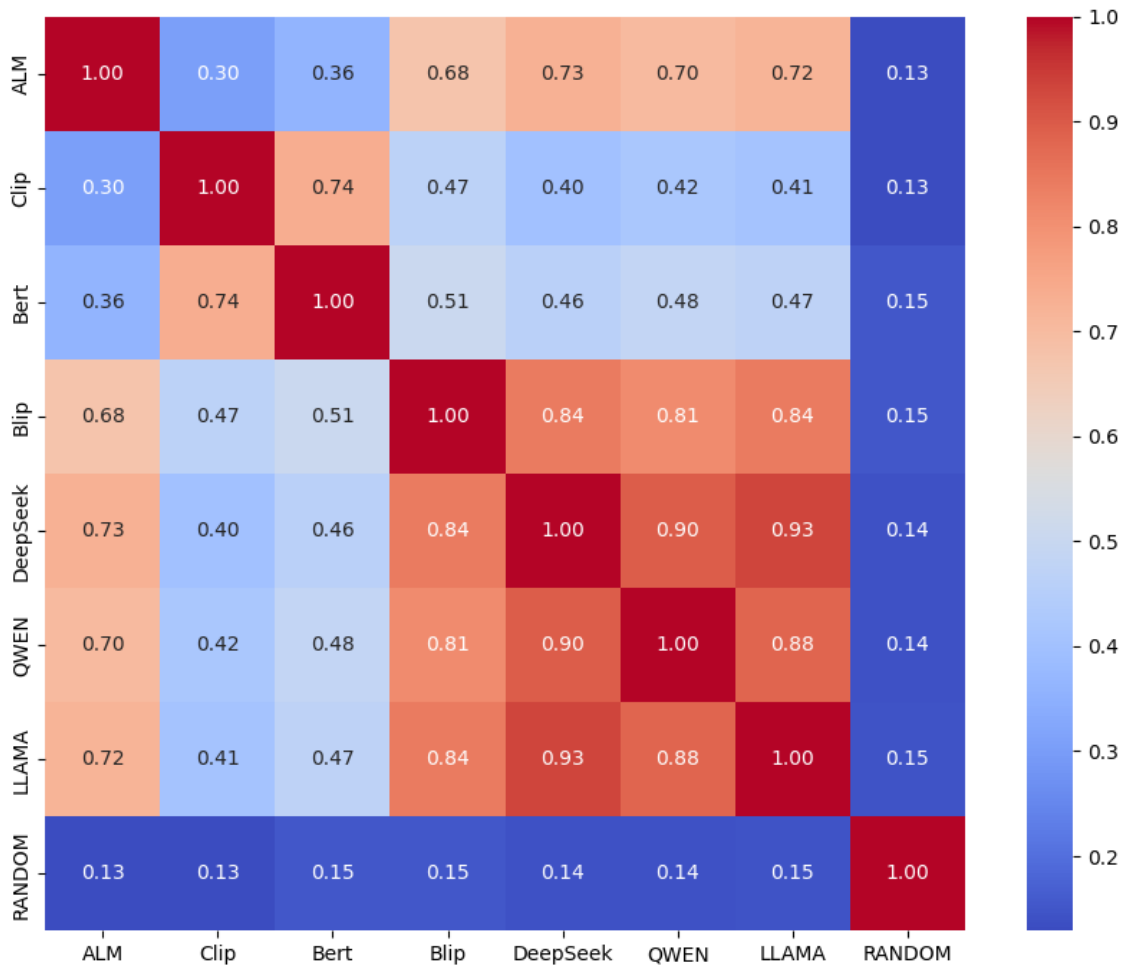

Figure 3. Heatmap illustrating the P@15 alignment between the vector spaces of the seven models considered and a randomly generated vector space (RANDOM). The models include our ALM, two VLMs (CLIP and BLIP) and four LLMs (BERT, Llama, BERT, Qwen and Deepseek). Results are averaged over 10 independent replications of the ALM experiment.

As shown, the alignment between the ALM and the decoder-only LLMs (Llama, Qwen and Deepseek) is relatively high, ranging from 0.70 to 0.73. A similarly strong alignment is observed with the BLIP VLM. In contrast, the alignment between the ALM and the CLIP VSL and the BERT encoder-only LLM is substantially lower. As expected, the alignment with the randomly generated vector space is minimal. The fact that this alignment does not approach zero reflects the nonzero probability that two sentences share a common neighbor within the top 15 out of 108 sentences (approximately 15%) purely by chance.

The highest alignment levels (from 0.83 to 0.93) are observed among the decoder-only LLMs which are trained on the same task. Notably, the alignment between the ALM and the decoder-only LLMs is only slightly lower, indicating a strong correspondence despite differences in training objectives and data.

Figure 4 and Table 1 present the P@k alignment between the ALM and the other models for different values of k [1, 5, 10, 15]. Across all values of k, the ALM exhibits consistently strong alignment with decoder-only LLM and BLIP, moderate alignment with CLIP and BERT, and negligible alignment with the random space. Remarkably, even in the P@1 condition, in which the two spaces must share the same nearest neighbor, the observed alignment is an order of magnitude higher than that of the random baseline.

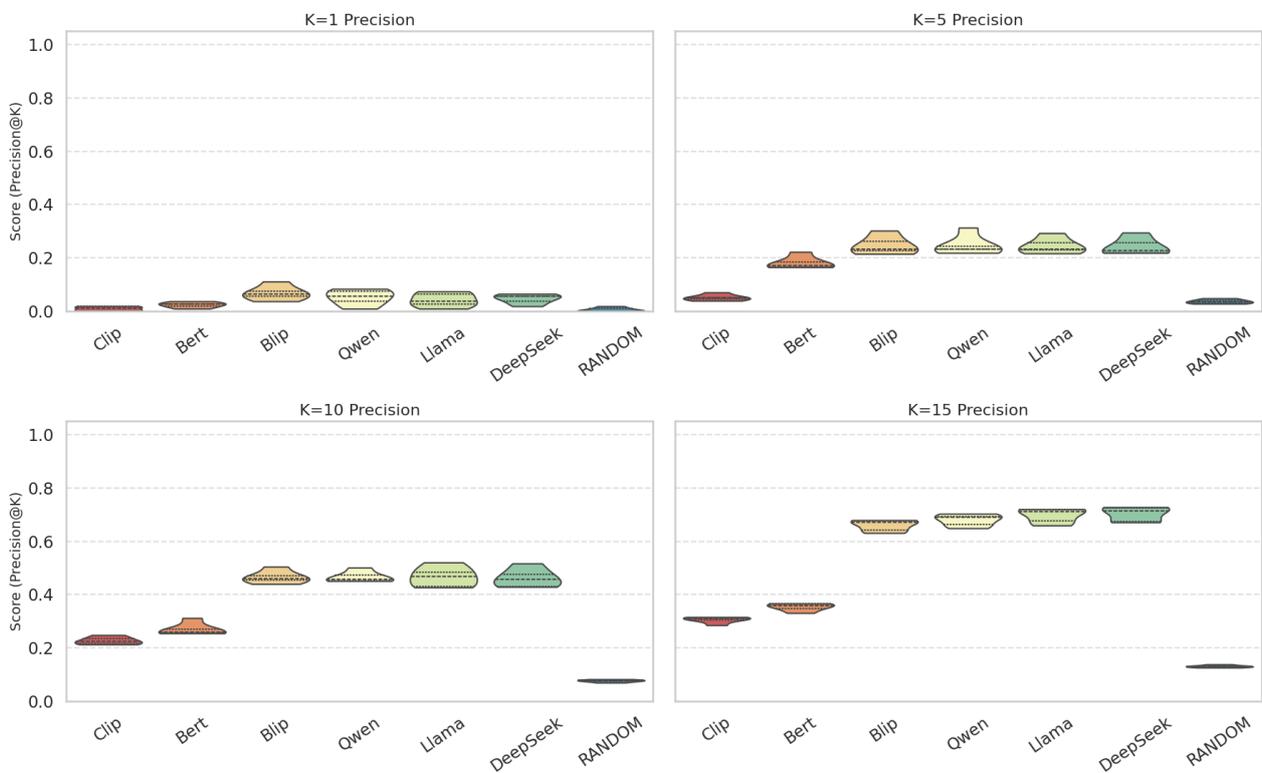

Figure 4. Alignment of the ALM vector space with the vector spaces of the other model considered and with a randomly generated vector space. The four figures show the Precision@K scores calculated with different k values [1, 5, 10, 15]. For each model, the data are aggregated across ten independent ALM initializations and training runs. The width of each violin plot represents the kernel probability density estimate of the scores, the internal lines indicate the interquartile range (25th to 75th percentile), and the horizontal marker denotes the median score.

| K | Clip | Bert | Blip | Qwen | Llama | DeepSeek | RANDOM |
|---|---|---|---|---|---|---|---|
| 1 | 0.0093 (0.0093) | 0.0241 (0.0106) | 0.0685 (0.0275) | 0.0519 (0.0297) | 0.0426 (0.0267) | 0.0463 (0.0185) | 0.0056 (0.0083) |
| 5 | 0.0515 (0.0119) | 0.1819 (0.0241) | 0.2474 (0.0349) | 0.2478 (0.0374) | 0.2456 (0.0298) | 0.2433 (0.0326) | 0.0367 (0.008) |
| 10 | 0.227 (0.0144) | 0.2704 (0.0241) | 0.4661 (0.0242) | 0.4665 (0.0211) | 0.4665 (0.0389) | 0.462 (0.0362) | 0.0767 (0.0048) |
| 15 | 0.3058 (0.0122) | 0.3536 (0.0146) | 0.6598 (0.0216) | 0.6793 (0.0225) | 0.6973 (0.0279) | 0.7022 (0.0281) | 0.1312 (0.0046) |

Table 1. Mean P@K, with standard deviation in brackets, between the ALM and all the other models at different level of K: 1, 5, 10 and 15.

To assess the robustness and significance of cross-model alignment, we conducted a nonparametric permutation test. Specifically, we randomly permuted the correspondence between sentences in one model's similarity matrix and recomputed P@15 over 1000 permutations. The empirical p-value was defined as the proportion of permuted scores greater than or equal to the observed alignment. Across all model combinations, except for the random embeddings, the observed alignments were significantly higher than chance ($p < 0.001$), confirming that the measured cross-model alignment is not due to random effects.

To further investigate the internal structure of the embedding spaces and quantify alignment, we performed a Procrustes analysis. As shown in Figure 5, the models exhibit a clear hierarchy of structural alignment. Qwen, LLaMA, and DeepSeek achieved the lowest median disparities (approximately 0.75–0.80), indicating that their representations exhibit the closest structural correspondence to the Behavioral Cloning embedding space after transformation. In contrast, BLIP and BERT showed the highest median disparities (all above 0.85 and approaching the random baseline of 1.0), suggesting that their embedding spaces are less readily aligned with the ALM structure. CLIP consistently displayed a high disparity, close to the random baseline, validating the analysis by yielding a score near 1.0.

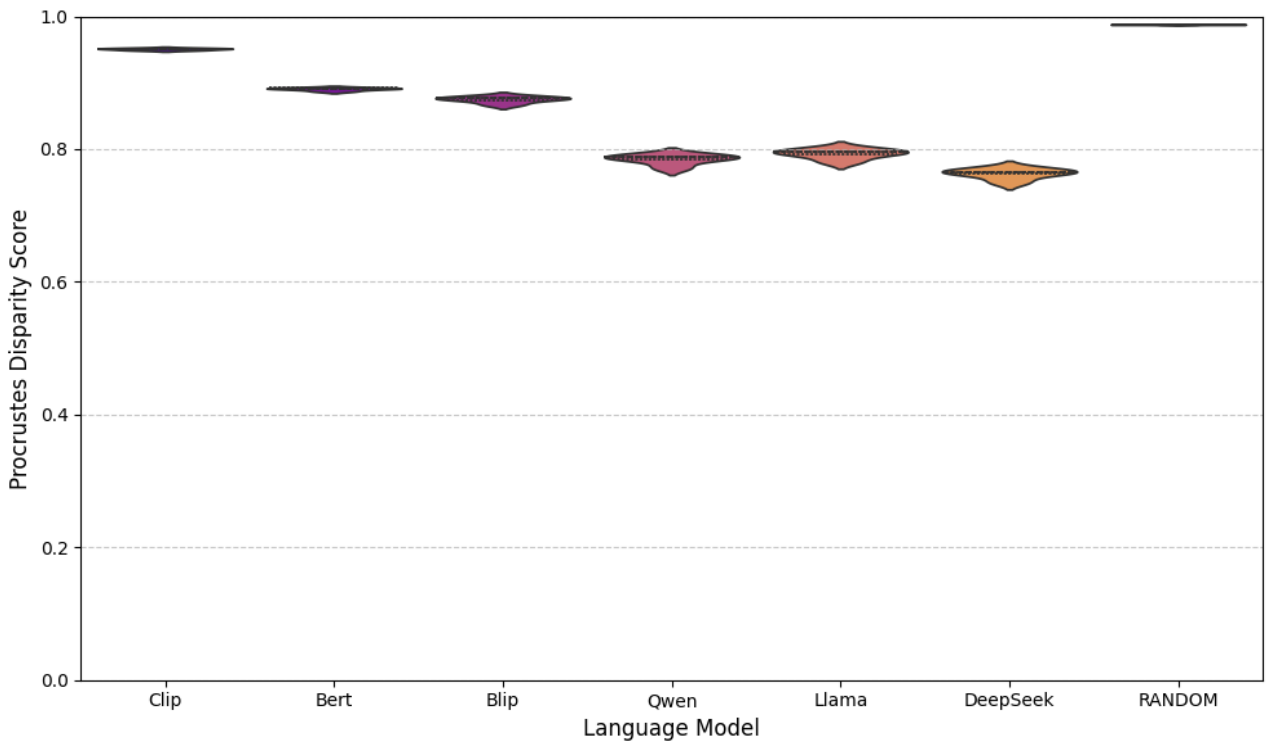

Figure 5. Violin plot illustrating the distribution of Procrustes disparity (residual error after optimal orthogonal transformation) for six language models compared with different

replications of the LAM model. Each violin represents ten disparity scores obtained by comparing a given model embedding space to the embedding space of ALM model. Lower disparity scores indicate greater geometric similarity between the models and ALM embedding spaces. The interior markers indicate the quartiles, with the white dot representing the median disparity. The width of each violin reflects the density of the scores, providing insight into the consistency and typical range of geometric similarity for each model across multiple ALM replications.

**4. Discussion**

The primary objective of this study was to investigate the degree of alignment among representations learned by large language models (LLMs), vision–language models (VLMs) and behavioral models trained to generate action sequences in response to linguistic commands. While prior research has explored the alignment between language and vision representations, the present work extends this inquiry to include the action dimension, thereby addressing whether representations supporting goal-directed behavior in situated agents share structural similarities with representations derived from passive linguistic or visual–linguistic learning.

To this end, we trained an agent through a behavioral cloning algorithm to map linguistic instructions and visual observations onto discrete motor actions. The embeddings produced by this model, called action embeddings were compared to those generated by several state-of-the-art LLMs (Llama, Qwen, DeepSeek, and BERT) and VLMs (CLIP and BLIP). Inter-model alignment was quantified by comparing the pairwise cosine similarity matrices of sentence embeddings and by evaluating the overlap in their nearest-neighbor structures through the precision-at-k (P@k) metric.

The results reveal a surprisingly high level of structural alignment among the representations derived from these three distinct modalities. Despite being trained on different data distributions and through distinct objectives—next-word prediction in LLMs, cross-modal contrastive learning in VLMs, and behavior imitation in the action model—the resulting embedding spaces exhibit substantial correlation, with P@k values ranging from 0.49 to 0.67 between the action model and the LLM/VLM spaces. Remarkably, the degree of alignment between the action model and certain LLMs (notably Llama and Qwen) is comparable to that observed among different LLMs themselves or among different VLMs. This finding suggests that models grounded in language, vision, and action converge toward partially shared representational geometries, despite their distinct training conditions and modalities.

This convergence is notable given the contrasting nature of the underlying learning processes. LLMs acquire linguistic knowledge from vast, text-only corpora through predictive modeling of symbol sequences, while VLMs learn through exposure to paired visual and textual data that emphasize perceptual–semantic correspondences. In contrast, the action model learns through direct interaction within a simulated environment, grounded in sensorimotor contingencies and reward-based imitation. Yet, the resulting embeddings organize linguistic instructions in a manner that mirrors the semantic structures encoded in LLM and VLM spaces. This finding indicates that representations supporting situated action and those supporting language and vision processing capture overlapping dimensions of meaning.

The high degree of alignment observed across models provides evidence that internal representations shaped by different modalities and tasks nonetheless embody comparable forms of semantic organization. This implies that the underlying structure of meaning may be modality-independent: linguistic, visual, and action knowledge can all converge on a shared representational substrate that supports transfer and generalization among modalities. In particular, the ability of an action model trained solely through behavioral imitation to encode linguistic instructions in a way that correlates strongly with LLM embeddings suggests that knowledge learned from interaction with the physical world can be integrated, or even anticipated, by models trained exclusively on language.

Such findings have important theoretical implications. They support the hypothesis that semantic representations grounded in linguistic experience alone may suffice to approximate, to some extent, the representational geometry required for sensorimotor reasoning. Conversely, action-grounded learning can give rise to representations that align naturally with linguistic and visual modalities, even in the absence of explicit multimodal training. This bidirectional alignment underscores the possibility of shared cognitive structures linking linguistic understanding, perceptual recognition, and goal-directed behavior.

From a practical standpoint, these results highlight the potential for cross-domain transfer: models trained in one modality may provide a foundation for rapid adaptation in another. For instance, LLMs could potentially support the initialization of language-conditioned control systems (Milano and Nolfi, 2025), while action-grounded models might enhance the grounding of language models in perceptual and motor semantics. The observed structural correspondence among modalities thus offers a pathway toward more integrated architectures capable of robust generalization across language, vision, and action domains.

While the present results provide clear evidence of structural alignment among language, vision, and action representations, several limitations should be acknowledged. First, the analysis is based on a restricted set of linguistically simple instructions drawn from a closed vocabulary within the BabyAI environment, which limits both linguistic diversity and semantic richness. Second, the action representations are learned in a single, highly simplified simulated environment characterized by discrete actions and constrained perceptual inputs. Finally, the alignment analyses rely on geometric similarity measures that capture relational structure but do not directly assess functional equivalence or causal grounding. Accordingly, the observed correspondences should be interpreted as evidence for shared representational organization rather than identical semantic grounding or understanding. Future work should investigate whether these alignment effects generalize to more complex environments, richer linguistic inputs, continuous action spaces, and alternative learning paradigms, thereby further clarifying the scope and limits of cross-modal representational convergence.

**Conclusion**

In summary, this study demonstrates that representations learned through fundamentally different learning processes—text-based prediction, visual–linguistic alignment, and situated action learning—nonetheless exhibit substantial structural similarity. The unexpectedly high degree of alignment among LLMs, VLMs, and the action model suggests that these systems, despite their differences in task objectives, training data, and degree of embodiment, converge toward shared representational structures that capture core aspects of semantic organization. This convergence supports the view that such representations form a bridge between language and action, facilitating knowledge transfer and generalization across modalities. Future work may explore whether this alignment extends to more complex tasks and environments, and whether explicitly leveraging these cross-modal correspondences can improve the grounding and adaptability of multimodal agents.